\documentclass[twoside,leqno,twocolumn]{article}
\newcommand{\mname}{HDUVA}
\newcommand{\fullalgoname}{Hierarchical Domain Unsupervised Variational Auto-encoding}
\newcommand{\lname}{LHDUVA}
\newcommand{\appname}{Appendix}
\usepackage[american]{babel}
\usepackage{natbib}
\usepackage{mathtools} 
\usepackage{booktabs} 
\usepackage[letterpaper]{geometry}
\usepackage{ltexpprt}
\usepackage{mathtools} 
\usepackage{booktabs} 
\usepackage{amsmath}
\usepackage{graphicx}
\usepackage{algorithm, algorithmic}
\usepackage{array}
\usepackage{subfig}
\usepackage{caption}
\usepackage{ifthen}
\usepackage{multicol} 
\usepackage[draft]{hyperref}

\usepackage{array}
\usepackage{booktabs}
\usepackage{threeparttablex}

\usepackage{graphicx}
\usepackage{capt-of}

\newcommand{\mcitep}{\citep}
\newcommand{\mcitet}{\citet}
\begin{document}
\title{\Large Hierarchical Variational Auto-Encoding for Unsupervised Domain Generalization\thanks{Presented at ICLR 2021, RobustML Workshop.}}
\author{Xudong Sun\thanks{LMU Munich, smilesun.east@gmail.com, 
part of this work was done during research internship at Siemens AG.}
\and Florian Buettner\thanks{Siemens AG, buettner.florian@siemens.com}}

\date{}

\maketitle







\begin{abstract} \small\baselineskip=9pt 
We address the task of domain generalization, where the goal is to train a predictive model such that it is able to generalize to a new, previously unseen domain. We choose a hierarchical generative approach within the framework of variational autoencoders and propose a domain-unsupervised algorithm that is able to generalize to new domains without domain supervision. We show that our method is able to learn representations that disentangle domain-specific information from class-label specific information even in complex settings where domain structure is not observed during training. Our interpretable method outperforms previously proposed generative algorithms for domain generalization as well as other non-generative state-of-the-art approaches in several hierarchical domain settings including sequential overlapped near continuous domain shift. 
It also achieves competitive performance  on the standard domain generalization benchmark dataset PACS compared to state-of-the-art approaches which rely on observing domain-specific information during training, as well as another domain unsupervised method. Additionally, we proposed model selection purely based on Evidence Lower Bound (ELBO) and also proposed weak domain supervision where implicit domain information can be added into the algorithm.

\end{abstract}
\section{Background and Motivation}
One big challenge of deploying a neural network model in real world use-cases is domain shift. In many real world applications, data seen by a deployed model is drawn from a distribution that is different from the training distribution and often unknown at train time. Domain Generalization aims at
training a model from a set of domains (i.e. related distributions) such that the model is able to generalize to a new, unseen domain at test time.
\ifx \flagworkshop \undefined
~Domain generalization is relevant for a variety of tasks, ranging from personalized medicine, where each patient corresponds to a domain, to predictive maintenance in the context of industrial AI. In the latter use-case, domains can represent different factories where an industrial asset (e.g. a tool machine or a turbine) is operated, or different workers operating the asset. In addition to these discrete domains,
\else
\fi 
~domain shift can manifest itself in a continuous manner, where for example the data distribution seen by an industrial asset can change due to wear and tear or due to maintenance procedures. Similarly, domain sub-structures are not always observable during training due to data privacy concerns (in particular when patient data is used). In these latter scenarios, it is difficult to train standard domain generalization algorithms since they are based on the notion of clearly separable domains that are observable during model training. 
In many of these use cases, interpretability and human oversight of machine learning models is key. Generative models allow for learning disentangled representations that correspond to specific and interpretable factors of variation, thereby facilitating transparent predictions.

We propose a new generative model that solves domain generalization problems in an interpretable manner without requiring domain labels during training. We build on previous work using autoencoder-based models for domain generalization \mcitep{kingma2013auto,ilse2019diva} and propose a \fullalgoname \,that we refer to as \mname.
Our major contributions include:
\begin{itemize}
\item We present an unsupervised algorithm for domain generalization that is able to learn in setting with incomplete or hierarchical domain information. Our algorithm only need to use extended ELBO as model selection criteria, instead of relying on the validation set. 
\item Our method is able to learn representations that disentangle domain-specific information from class-label specific information without domain supervision even in complex settings.
\item Our algorithm generates interpretable domain predictions that reveal connections between domains.
\item We constructed several hierarchical and sequential domain generalization benchmark datasets with doubly colored mnist for the domain generalization community.
\item Our method allows weak domain supervision by adding partially observed domain information into the algorithm.
\end{itemize}


\section{Related work}\label{sec:relate}
In this section, we provide a taxonomy of existing solutions in domain generalization.
In general, domain generalisation approaches can be divided into the following main categories, that we describe below. 
\paragraph{Invariant Feature Learning}\label{para:ifl}
While observations from different domains follow different distributions, Invariant Feature Learning approaches try to map the observations from different domains into a common feature space, where domain information is minimized \mcitep{xie2017controllable,akuzawa2018domain}. The method works in a mini-max game fashion in that there is a domain classifier trying to classify domains from the common feature space, while a feature extractor tries to fool this domain classifier and help the target label classifier to classify class label correctly. 
\mcitet{li2017deeper} presented a related approach and used tensor decomposition to learn a low rank embedding for a set of domain specific models as well as a base model. We classify this method into invariant feature learning because the base model is domain-invariant.
\paragraph{Image Processing Based Method}\label{para:improcess} \mcitet{carlucci2019domain} divided the image into small patches and generated permutations of those small patches. They then used a deep classifier to predict the predefined permutation index so that the model learned the global structure of an image instead of local textures.
\mcitet{wang2019learning} used a gray level co-occurence matrix to extract superficial statistics. They presented two methods to encourage the model to ignore the superficial statistics and thereby learn robust representations. This group of methods has been developed for image classification tasks, and it is not clear how it can be extended to other data types. 
\paragraph{Adversarial Training Based Data Augmentation}\label{para:adverse_aug}
\mcitet{Volpi2019} optimized a procedure to search for worst case adversarial examples to augment the training domain. \mcitet{volpi2018generalizing} used Wasserstein distance to infer adversarial images that were close to the current training domain, and trained an ensemble of models with different search radius in terms of Wasserstein distance.

\paragraph{Meta Learning Based Method\label{para:meta}} Meta learning based domain generalization method (MLDG) uses model agnostic training to tackle domain generalization as a zero-shot problem,  by creating virtual train and test domains and letting the meta-optimizer choose a model with good performance on both virtual train and virtual test domains \mcitep{li2018learning}. \mcitet{balaji2018metareg} improved upon MLDG by concatenating a fixed feature network with task specific networks. They parameterized a learnable regularizer with a neural network and trained with a meta-train and a meta-test set .
\paragraph{Auto-Encoder Based Method}\label{para:ae_method_dg} DIVA \mcitep{ilse2019diva} builds on variational auto-encoders and splits the latent representation into three latent variables capturing different sources of variation, namely class specific information ($z_y$), domain specific information ($z_d$) and residual variance ($z_x$). Disentanglement is encouraged via conditional priors, where the domain-specific latent variable $z_d$ is condition on an observed, one-hot-encoded domain $d$.  As auxiliary components, DIVA adds a domain classifier based on $z_d$, as well as a target class label classifier based on $z_y$. 
\mcitet{hou2018cross} encoded images from different domains in a common content latent code and domain-specific latent code, while the two types of encoders share layers. Corresponding discriminators are used to predict whether the input is drawn from a prior distribution or generated from encoder.
\paragraph{Causality based Method}
Recently, \mcitet{mahajan2020domain} proposed MatchDG with that approximates base object similarity by using a contrastive loss formulation adapted for multiple domains. The algorithm then match inputs that are similar under the invariant representation. \\

Comparing these families of approaches, we can see that only probabilistic auto-encoder based models inherit advantageous properties like semi-supervised learning, density estimation and variance decomposition naturally. While autoencoder-based approaches such as DIVA have a better interpretability than all other approaches, a main drawback is that explicit domain labels are required during training.
This can be problematic in a number of settings. In particular,  a one-hot encoding of domains does not reflect scenarios where a continuous domain shift can occur. In this case, without knowledge of the causal factor that causes the domain shift, it is not clear how such continuous shifts can be one-hot encoded in a meaningful manner. In addition, 

\begin{itemize}
\item Domains can have a hierarchical structure reflected by related sub-domains (e.g. country $>$ factory $>$ machine). One-hot encodings as used in existing autoencoder-based approaches are not able to model such hierarchical domain structures.
\item In some applications, domains are not necessarily well-separated, but significant overlap between domains can occur (e.g. a cartoon might look more similar to a  pop-art painting than a photography). One-hot encoding such overlapping domains encourages separated representations, which may harm model performance.
\item A one-hot encoding of domains mapping to the prior distribution of $z_d$ may limit the generalization power of neural networks, especially when we deal with continuous domain shift. 
\end{itemize}
\begin{figure}[h!]
\centering
	\subfloat[Instance View]{
	\includegraphics[width=0.2\textwidth]{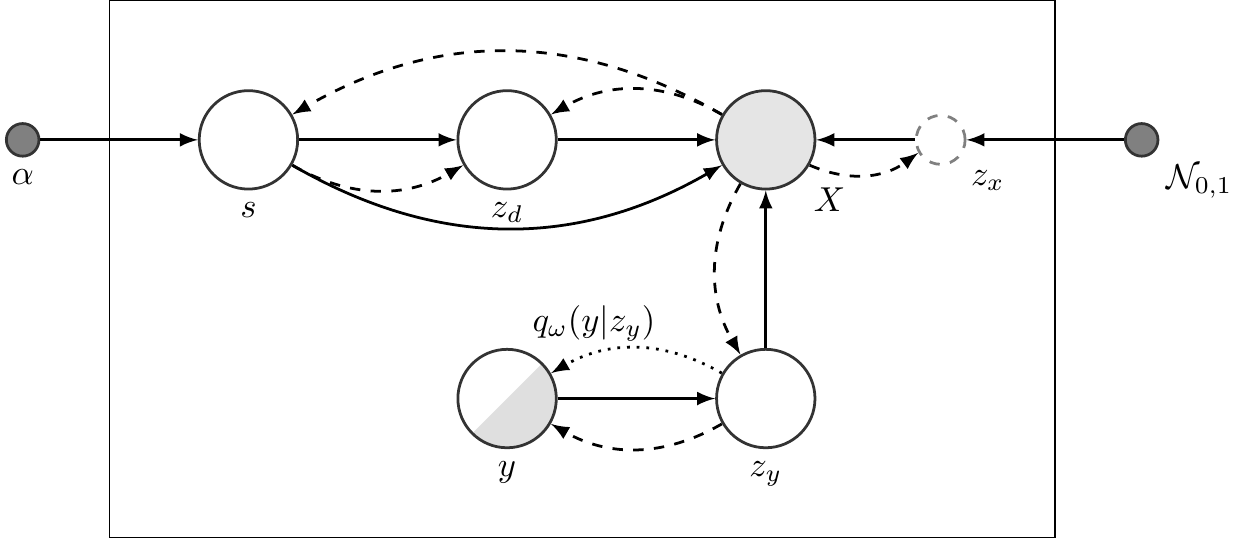}\label{fig:hdiva_instance_view}}
	\subfloat[Domain Batch View]{
	\includegraphics[width=0.2\textwidth]{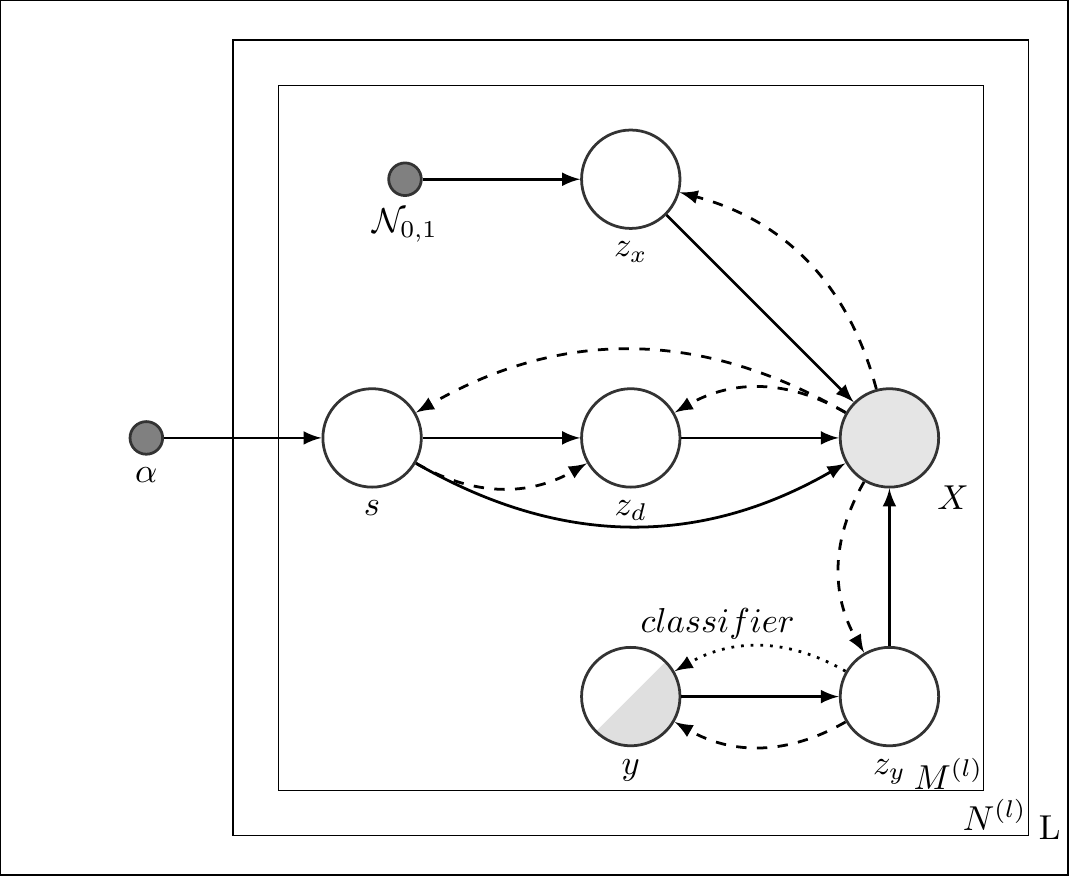}\label{fig:ehdiva}}
	\caption{\textbf{\mname: \fullalgoname}}
	\label{fig:ehdiva-two}
\end{figure}
\section{Methods and Technical Solution}\label{sec:method}
\subsection{Problem statement and notation}
Domain generalization aims to generalize models to unseen domains without knowledge about the target distribution during training. A domain $d$ consists of a joint distribution $p(x,y)$ on $\mathcal{X}\times \mathcal{Y}$, with $\mathcal{X}$ being the input space and $\mathcal{Y}$ being the output space \mcitep{muandet2013domain}. For our modelling approach, we employ the framework of variational autoencoders (VAEs) \mcitep{kingma2013auto}. We use $z$ to represent the latent representation of a VAE and use three independent latent representations to disentangle variability in inputs $X$ related to domain-specific sources, label-specific sources and residual variation. We use probabilistic graphical  models to illustrate the conditional dependecies of random variables, observables and hyperparameters in Figure \ref{fig:ehdiva-two}. In the graphical model of Figure \ref{fig:ehdiva-two}, solid circles  represent observations and white circles represent latent variables. We use half-shaded circles to represent a variable can either be observed or act as latent variable, which is typical in semi-supervised learning. Small solid circles in Figure \ref{fig:ehdiva-two} represent fixed  hyper-parameters. Subscripts represent  components of a variable, while we use super-script to index samples and domains. We use solid arrows to represent generative path, and dashed arrows to represent variational inference part. Plates represent repetitions of random variables. We use $\theta$ to represent learnable parameters of priors/decoders and $\phi$ to represent learnable parameters of variational posterior distributions/encoders.
\subsection{\mname{} overview}\label{sec:hdiva}
To overcome the limitations of current autoencoder-based methods, we propose a hierarchical probabilistic graphical model called  \fullalgoname, in short \mname{}. Our model is based on three latent variables to model distinct sources of variation that are denoted as $z_y$, $z_d$ and $z_x$. $z_y$ represents class specific information, $z_d$ represents domain specific information and $z_x$ models residual variance of the input. For model ablation on $z_x$, we also developed a version of \mname{} without $z_x$ and evaluate the effect empirically in Section~\ref{sec:exp}. In Figure \ref{fig:hdiva_instance_view}, we use dashed node to denote $z_x$ to indicate that $z_x$ and its associated components in the model can be removed which result in Figure~\ref{fig:hduva_no_zx} in \appname~\ref{sec:no-zx}.  We model the prior distribution of $z_y$ as conditional distribution based on class label $y$. These settings together with the auxiliary classifier in Equation~\ref{eq:ehdiva_elbo_plus_y} constitutes the inductive bias of the model as promoted in \mcitet{locatello2019challenging}. We introduce an additional  hierarchical level and use a continuous latent representation $s$ to model (potentially unobserved) domain structure. This means that we can encourage disentanglement of the latent variables through conditional priors without the need of conditioning on a one-hot-encoded, observed domain label as in \mcitet{ilse2019diva}.\\
More specifically, we first place a Dirichlet prior on $s$ such that it can be interpreted as a soft, topic-like, version of the standard one-hot encoded domain $d$. 
We then use $z_d$ to capture domain-specific variation by conditioning its prior on $s$. Note that in our model this domain $s$ is not an observable but instead a latent variable to be inferred from data. For clarity, we refer to an observed domain as nominal domain.  Borrowing from topic models in NLP \mcitep{srivastava2017autoencoding, pmlr-v119-gupta20a}, we refer to $s$ as topic. 
In Figure~\ref{fig:hdiva_instance_view}, we form a hierarchical path \mcitep{klushyn2019learning} from topic $s$ to $z_d$ to observation $x$. 
We use $K$ to denote the dimension of the domain representation or topic vector $s$, i.e. $dim(s) = K$. We use $k$ to index each component of $s$, i.e. $s^{(l)} = [s^{(l)}_1, s^{(l)}_2, \cdots, s^{(l)}_K]$, with $l$ indexing a domain.  Note that in our case, $K$ can be either greater, smaller or equal to the number of domains $L$, while in supervised approaches, the one-hot encoded domain label is always the size of $L$. This is beneficial not only in setting with unobserved domain observation, but also for problems with a large number of domains which lie on a lower-dimensional manifold (e.g. thousands assets in an predictive maintenance task). In practice, we set $K=3$ for easy visualization, see Figure~\ref{fig:topic_mnist_color}. 
Figure \ref{fig:ehdiva} offers a domain-batch view of \mname{} for stochastic gradient descent, where the batch size is denoted by $M^{(l)}$ for the $l$th domain,  with a total of $N^{(l)}$ batches for domain $l$. We use $i$ to index a batch and $j$ to index a sample. For simplicity, $i$ and $j$ are omitted whenever convenient and not causing confusion. The domain-batch view for our model can be useful in \appname~\ref{sec:weak} when we talk about weak domain supervision for incorporating partially observed domain information into the model.
\ifx \flagworkshop \undefined
\else
Details on model implementation and inference are described in the supplement. 
\fi
Taken together, we present a novel approach for probabilistic domain generalization without the need for observed domain labels. 
\ifx \flagworkshop \undefined
  
\else
Further details about the prior setting and inference process for our method can be found in the supplement material.
\fi

\subsection{Model implementation}
In this section, we first describe the generative model with prior distributions, followed by a discussion on model inference.
\subsubsection{Prior Distributions for $z_x$, $z_y$ and $z_d$}
We chose a standard isotropic Gaussian prior with zero mean and unit variance for $z_x$ and conditional priors for for $z_y$ and $z_d$. More specifically, we chose a normal prior for $z_y$ that is conditioned on the target class label $y$:
\begin{align}
&  p_{\theta_y}(z^{(l,i)}_{y} |y^{(l,i)}) = \mathcal{N}\left(\cdot|\mu_{\theta_y}(y^{(l,i)}), \sigma_{\theta_y}(y^{(l,i)})\right)\label{eq:prior_zy}
\end{align}
with $\mu_{\theta_y}$ and $\sigma_{\theta_y}$ being learnable parameterizations of the mean and standard deviation in form of neural networks. Similarly, we choose a normal prior for $z_d$ and condition it on $s$:
\begin{align}
&  p_{\theta_d}(z^{(l,i)}_{d} |s^{(l,i)}) = \mathcal{N}\left(\cdot|\mu_{\theta_d}(s^{(l,i)}), \sigma_{\theta_d}(s^{(l,i)})\right)\label{eq:prior_zd}
\end{align}
where again $\mu_{\theta_d}$ and $\sigma_{\theta_d}$ parameterize mean and variance of $z_d$.
\subsubsection{Prior Distribution for $s$}
We would like for $s$ to display topic-like characteristics, facilitating interpretable domain representations. Consequently, we use a Dirichlet prior on $s$, which is a natural prior for topic modeling \mcitep{srivastava2017autoencoding,joo2020dirichlet,Zhao_Wang_Masoomi_Dy_2019}. 

Let $\alpha$ be the Dirichlet concentration parameter $\alpha = [\alpha_1, \alpha_2, \cdots, \alpha_K]$, then the prior distribution of $s$ can be written as: 
\begin{equation}
p(s^{(l,i)}|\alpha^{l}) = Dir(s^{(l,i)}|\alpha_{1:K}^{l}) = \frac{\prod_k (s^{(l,i)}_k)^{\alpha^{l}_k - 1}}{\mathcal{Z}(\alpha^{l}_{1:K})}\label{eq:dirichlet_prior}
\end{equation}
where we use $\mathcal{Z}(\alpha_{1:K})$ to represent the partition function.


We do not learn the distribution parameter $\alpha$, but instead, leave it as a hyper-parameter. By default, we set $\alpha$ to be a vector of ones, which corresponds to a uniform distribution of topics. We refer to this prior setting as flat prior for unsupervised domain generalization. If more prior knowledge about the relation between training domains is available, an informative prior can be used instead. Additionally, we provide further techniques to add partially observed domain information into a weak domain supervision fashion as explained in detail in \appname~\ref{sec:weak}.
\subsubsection{Inference for \mname{}} 
We perform variational inference and introduce three separate encoders as follows, where the hierarchical inference on $z_d$ and $s$ follows \cite{tomczak2018vae}. In \appname~\ref{sec:lhdiva}, we offer an alternative hierarchial inference method following Ladder VAE in \cite{sonderby2016ladder} which we term L\mname{} in the experiment section.
\begin{align}
&q_{\phi}(s^{(l,i)}, z_d^{(l,i)}, z_x^{(l,i)}, z_y^{(l,i)}|x^{(l,i)})\nonumber\\
=&q_{\phi_s}(s^{(l,i)}|x^{(l,i)}) 
q_{\phi_d}(z_d^{(l,i)}|s^{(l,i)}, x^{(l,i)})
q_{\phi}( z_x^{(l,i)}, z_y^{(l,i)}|x^{(l,i)})  \label{eq:qs_qzd_qzx_qzy}
\end{align}
For the approximate posterior distributions of $z_x$ and $z_y$, we assume fully factorized Gaussians with parameters given as a function of their input:
\begin{align}
q_{\phi}(z_x^{(l,i)}, z_y^{(l,i)}|x^{(l,i)})
=q_{\phi_x}(z_x^{(l,i)}|x^{(l,i)})q_{\phi_y}(z_y^{(l,i)}|x^{(l,i)})  \label{eq:qzd_qzx_qzy}   
\end{align}
Encoders $q_{\phi_s}$ , $q_{\phi_d}$, $q_{\phi_y}$, and $q_{\phi_x}$ are parameterized by $\phi_s$, $\phi_d$, $\phi_y$, and $\phi_x$ using separate neural networks to model respective means and variances as function of $x$.



For the form of the approximate posterior distribution of the topic $s$ we chose a Dirichlet distribution:
\begin{align}
&  q_{\phi_s}(s^{(l,i)}|x^{(l,i)})
= Dir\left(s^{(l,i)}| \phi_{s}(x^{(l,i)})\right)\label{eq:ehdiva_alpha_infer_topic}\end{align}
where $\phi_{s}$ parameterizes the concentration parameter based on $x$, using a neural network. We use the technique in \mcitet{jankowiak2018pathwise} to reparameterize the Dirichlet distribution.
\subsubsection{ELBO for \mname{}}
Given the priors and factorization described above, we can optimize the model parameters by maximizing the evidence lower bound (ELBO). 
We can write the ELBO for a given input-output tupel $(x,y)$ as:
\begin{align}
&ELBO(x,y) = E_{q(z_d, s|x), q(z_x|x), q(z_y|x)}\log p_{\theta}(x|s, z_d, z_x, z_y) \nonumber\\
&- \beta_x KL(q_{\phi_x}(z_x|x)||p_{\theta_x}(z_x)) - \beta_y KL(q_{\phi_y}(z_y|x)||p_{\theta_y}\nonumber\\
&(z_y|y)) - \beta_d 
E_{q_{\phi_s}(s|x), q_{\phi_d}(z_d|x, s)} \log \frac{q_{\phi_d}(z_d|x, s)}{p_{\theta_d}(z_d|s)} \nonumber\\ &-\beta_s E_{q_{\phi_s}(s|x)}KL(q_{\phi_s}(s|x)||p_{\theta_s}(s|\alpha)) \label{eq:ehdiva_elbo}
\end{align}
where we use $\beta$ to represent the multiplier in the Beta-VAE setting \mcitep{higgins2016beta}, further encouraging disentanglement of the latent representations. Note that we do not consider $\beta$ to be hyper-parameters with respect to the performance of the task (e.g. classification), instead we set all $\beta$s to $1.0$ except otherwise stated.

Finally, we add an auxiliary classifier $q_{\omega}(y|z)$, which is parameterized by $\omega$, to encourage separation of classes $y$ in $z_y$. The \mname{} objective then becomes:
\begin{align}
\mathcal{F}(x,y) = ELBO(x,y) + \gamma_y E_{q_{\phi_y}(z_y|x)}[\log q_{\omega}(y|z_y)]\label{eq:ehdiva_elbo_plus_y}
\end{align}
The whole process is described in Algorithm \ref{algo:ehdiva}. The objective function in Equation \ref{eq:ehdiva_elbo_plus_y} which we coin extended ELBO can also be used as a model selection criteria, thus our method does not need validation set at all,  as we empirically evaluated in the experimental section in section \ref{sec:exp}.  
\begin{algorithm}
	\caption{\mname{}} 
        \begin{algorithmic}[1] 
		\WHILE{not converged or maximum epochs not reached}
		\STATE warm up $\beta$ defined in Equation \ref{eq:ehdiva_elbo}, as in \mcitet{sonderby2016ladder}
		\STATE fetch mini-batch \{x, y\} =\{$x^{(l,i)},y^{(l,i)}$\}  \\
        \STATE compute parameters for $q_{\phi_x}(z_x|x)$, $q_{\phi_y}(z_y|x)$, $q_{\phi_s}(s|x)$, $q_{\phi_d}(z_d|s, x)$ \\
		\STATE sample latent variable $z_{x}^q$, $z_{y}^q$, $s^q$, $z_d^q$ and compute $[\log q_{\omega}(y|z_y)]$.
		\STATE compute prior distribution for $z_d$ using $s$
		\STATE compute $p_{\theta}(x|z_x,z_y,z_d, s)$ using sampled $s$, $z_{x}^q$, $z_{y}^q$, $z_{d}^q$
		\STATE compute KL divergence for $z_d$, $z_x$ and $z_y$, $s$.
		\STATE aggregate loss according to Equation \ref{eq:ehdiva_elbo_plus_y} and update model
		\ENDWHILE 
	\end{algorithmic}
	\label{algo:ehdiva}
\end{algorithm}


\section{Empirical Evaluation}\label{sec:exp}
We conduct experiments, trying to answer the following questions:
\begin{itemize}
\item Generative domain generalization approaches facilitate model interpretability with domain-concept disentanglement. However, in complex scenarios with domain substructure (hierarchical domains), compared to other generative approach for domain generalization, can \mname{} still robustly disentangle domain-specific variation from class-label specific variation? See details in Section~\ref{sec:color-combo}.
\item Compared to clear domain shift, hierarchical, near continuous and overlapped domain shift has been less well studied in the domain generalization community. How well does \mname{} perform under this situation compared to other domain generalization approaches. See Section~\ref{sec:cont_overlap}.
\item In medical image classification, images from each patient can form a separate domain. In complex settings where image source can only be associated to a group of patients, how does \mname{} perform compared to other methods? See Section~\ref{sec:malaria}. 
\item  Could \mname{} be used as a domain embedding tool? We visualize topics from colored mnist domains to illustrate the possiblity in Section~\ref{sec:topic_plot}.
\item 
How does \mname{} perform under standard domain generalization benchmarks where information on clearly separated domain is available, compared with other state-of-the-art algorithms? See Section~\ref{sec:pacs}.
\item How does the hyper-parameter $\gamma_y$ in Equation~\ref{eq:ehdiva_elbo_plus_y} affect the performance of our algorithm? In face of a new scenario, is there a way to help us to figure out a value for this hyper-parameter?  See Section~\ref{subsec:hyper-sensitivity}.
\end{itemize}
To make fair comparisons, we separate the implementation of each algorithm being compared from the task scenario they have to be tackle with thus enabling the same condition for each experiment repetition with a random seed. All algorithms share the same neural network architecture and we use recommended hyper-parameters for the competitor algorithm following the orginal literature.
A detailed experimental setting, including the architecture of the neural network used in the experiment, as well hyper-parameter for each algorithm, can be found in \appname~\ref{supsec:hyper_exp_set}.

\subsection{Generative Domain Generalization approches on Hierarchical Domains}\label{sec:color-combo}
\begin{figure}[h!]
\centering
\subfloat[1st domain]{\includegraphics[width=0.15\textwidth]
{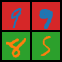}\label{fig:combo0}}
\subfloat[2nd domain]{\includegraphics[width=0.15\textwidth]{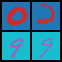}\label{fig:combo1}}
\subfloat[3rd domain]{\includegraphics[width=0.15\textwidth]{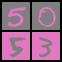}\label{fig:combo2}}
	\caption{\textbf{Random combination of Color-Mnist as Hierarchical Domains}. Mnist has both its foreground and background colored, each color combination represent one sub-domain. Each nominal domains include 2 sub-domains.}\label{fig:color_mnist_both_combo}
\end{figure}

To simulate domains with sub-structures (hierarchical domains), we create sub-domains within nominal domains, where all sub-domains within one nominal domain share the same domain label. We adapt color-mnist \mcitep{metz2016unrolled, rezende2018taming} with the modification that both its foreground (the figure itself) and background are colored as the sub-domain, as shown in  Figure \ref{fig:color_mnist_both_combo}. We constructed 3 nominal domains with sub-structures as indicated in Figure \ref{fig:color_mnist_both_combo}. For baseline algorithms which require a domain label during training, we use a one-hot encoded nominal domain as the explicit domain label. For \mname{}, we do not use domain label information. Besides, we only use extended ELBO in Equation \ref{eq:ehdiva_elbo_plus_y} as model selection criteria instead of a using a separate validation set while keeping the training set for all algorithms the same at each experiment repetition. Further experimental details can be found in \appname~\ref{supsec:hyper_exp_set}.

We are interested in evaluating how our domain-unsupervised approach and domain-supervised generative domain generalization algorithms like DIVA \cite{ilse2019diva} for domain generalization would behave under this sub-domain scenario, in terms of out-of-domain prediction accuracy and disentanglement performance. 
We perform a leave-one-domain-out evaluation \mcitep{li2017deeper}, where each test domain is repeated 10 times with 10 different random seeds. We report the out of domain test accuracy in Table \ref{tb:mnist-color-rand-combo}. Table \ref{tb:mnist-color-rand-combo} shows that \mname{} outperforms DIVA in terms of out of domain performance on all three test domains, while retaining a very small variance compared to DIVA.

\begin{table*}[t]
	\centering
	\caption{\textbf{Out of Domain Accuracy on Color-Mnist Composed Subdomain inside Nominal Domains}}
	\begin{tabular}{ lccc }
		\toprule
		 \bfseries Color-Mnist (Figure \ref{fig:color_mnist_both_combo}) & \bfseries Test Domain 1& \bfseries Test Domain 2 & \bfseries Test Domain 3  \\
		\midrule
		\mname{}  &\textbf{0.93} $\pm$ 0.02   &\textbf{0.69} $\pm$ 0.12    & \textbf{0.55} $\pm$ 0.03  \\
		
	DIVA \mcitep{ilse2019diva}  &0.88 $\pm$ 0.05    &0.56 $\pm$ 0.19 & 0.50 $\pm$ 0.08 \\ 
		
		
		\bottomrule
	\end{tabular}
	
	\label{tb:mnist-color-rand-combo}
\end{table*}

To explain such a performance difference, we further evaluate how robustly DIVA and \mname{} are able to disentangle different sources of variation under this scenario with incomplete sub-domain information.

\begin{figure}[!t]
\centering
\subfloat[DIVA]{\includegraphics[width=1.5in]
{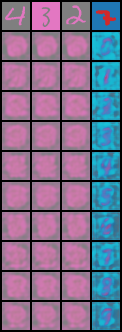}\label{fig:diva_0357}}
\subfloat[\mname{}]{\includegraphics[width=1.5in]{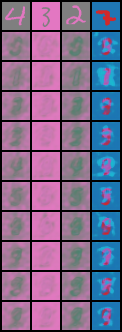}\label{fig:hdiva_0357}}
\caption{ \textbf{Comparison of Conditional Image Generation under Incomplete Domain Knowledge}. The domain composition is shown in Figure \ref{fig:color_mnist_both_combo}.}
\label{fig:fig_disentangle_combo_color}
\end{figure}

We sample seed images from different sub-domains as shown in the first row of Figure \ref{fig:fig_disentangle_combo_color}. We then generate new images by scanning the class label from 0 to 9 by sampling from the conditional prior distribution of $z_y$ (i.e.  $p_{\theta_y}(z_{y} |y)$, eq. \ref{eq:prior_zy}). We keep the domain representation the same as in the seed image, set the noise component $z_x$ to zero and then use the decoder network $p_{\theta}(x|z_d, z_x, z_y)$ to generate an image based on the three latent representations. 
If the models are able to disentangle domain-specific variation from class-label specific variation in $z_y$ and $z_d$, we expect that the generated images have the same domain information as the seed image (foreground and background color) while generating different class labels (numbers from 0 to 9). In Figure \ref{fig:fig_disentangle_combo_color} we compare DIVA and \mname{}'s generative performance. Due to the sub-structure inside the nominal domains, DIVA could only reconstruct a blur of colors for the first 3 columns in Figure \ref{fig:diva_0357}, while \mname{} could generate different numbers for 2 of the three seed images. For the last seed image, both DIVA and \mname{} could conditionally generate numbers, but DIVA did not retain the domain information (since the background color, which is dark blue in the seed image, is light blue in the generated images and the pink background color from the seed image becomes grey in the generated images). This indicates that DIVA is not able to disentangle the different sources of variation and domain information is captured by $z_y$ as well. In contrast, \mname{} was able to separate domain information from class-label information.


\subsection{Generalization under hierarchical, near continuous and overlapped domain-drift scenarios}\label{sec:cont_overlap} 
In some occasions, the boundaries between different domains can be ambiguous. For example, consider continuous domain drift in industry applications, some physical parameters of the same type of machine in different factories might change continuously and between two factories there can be overlap.

\begin{figure}[!t]
	\centering
		\subfloat[1st domain]{\includegraphics[width=0.15\textwidth]
			{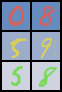}\label{fig:vlag-combo0}}
		\subfloat[2nd domain]{\includegraphics[width=0.15\textwidth]{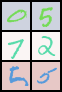}\label{fig:vlag-combo1}}
		\subfloat[3rd domain]{\includegraphics[width=0.15\textwidth]{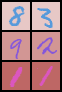}\label{fig:vlag-combo2}}
	\caption{\textbf{Sequential Color-Mnist (VLAG Palette)}. Background color taking 7 hue values spanning the \texttt{VLAG} hue ranges sequentially, with fixed saturation and lightning, foreground color takes equally spaced hue value in the complete hue circle with fixed saturation and lightning. Background and foreground colors are zipped. The first 3 color schemes representing 3 sub-domains compose the first nominal domain in Figure \ref{fig:vlag-combo0}, the 2nd nominal domain in Figure \ref{fig:vlag-combo1} takes the middle 3 color schemes with one color scheme overlap with the 1st and one color scheme overlap with 3rd nominal domain in Figure \ref{fig:vlag-combo2}. The 2nd nominal domain serves as a bridge between the other two nominal domains. Out of domain test accuracy is reported in Table \ref{tb:vlag-seq_color_overlap}.}
	\label{fig:vlag-color-mnist-cont-overlap}
\end{figure}

To simulate such a behavior, we consider a domain-drift scenario with Color-Mnist in Figure \ref{fig:vlag-color-mnist-cont-overlap}. By dividing a smooth color palette into 7 sub-domains (with each color corresponding to a sub-domain), we simulate a near continuous domain shift. We use this scenario to evaluate how robust our algorithm is in domain drift scenarios. Detailed construction process and experiment setting can be found in the figure caption.

\begin{figure}[!t]
	\centering
		\subfloat[1st domain]{\includegraphics[width=0.15\textwidth]
			{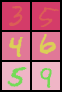}\label{fig:seq_overlap_combo0}}
		\subfloat[2nd domain]{\includegraphics[width=0.15\textwidth]{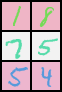}\label{fig:seq_overlap_combo1}}
		\subfloat[3rd domain]{\includegraphics[width=0.15\textwidth]{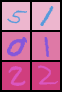}\label{fig:seq_overlap_combo2}}
	\caption{\textbf{Sequential Color-Mnist (Red Diverging Palette)}.
	Background color taking 7 hue values spanning in the area between 0 and 350 hue degrees (red spectrum) sequentially, with fixed saturation and lightning, foreground color takes equally spaced hue value in the complete hue circle with fixed saturation and lightning. Background and foreground colors are zipped. The first 3 color schemes representing 3 sub-domains compose the first nominal domain in Fig. \ref{fig:seq_overlap_combo0}, the 2nd nominal domain take the middle 3 color schemes with one color scheme overlap with the 1st and another color scheme overlap with the 3rd nominal domain in Fig. \ref{fig:seq_overlap_combo1}. The 2nd nominal domain serves as a bridge between the other two nominal domains. Out of domain test accuracy is reported in Table \ref{tb:red-diverging-seq_color_overlap}.}
	\label{fig:red-diverging-color-mnist-seq-overlap}
\end{figure}

Following  leave-one-domain out setting as in other experiments, we report the out-of-domain classification accuracy in Table \ref{tb:vlag-seq_color_overlap}, illustrating that our unsupervised approach is better able to account for continuous domain drift scenarios than standard supervised approaches that artificially categorize the gradually shifting into distinct nominal domains. To remove the effects of the choice of the color palette on the results, we conduct similar experiments on another diverging color palette as shown in Figure
\ref{fig:red-diverging-color-mnist-seq-overlap} and the corresponding results are listed in Table \ref{tb:red-diverging-seq_color_overlap} where the detailed scenario construction process and experimental setting are explained in the caption of the figure and table. From these results, we find that \mname{} has significant comparative advantage on the constructed hierarchical, continuous and overlapped domain shift scenarios over DIVA, MatchDG and the Deep-All baselines.

\begin{table}[!t]
	\centering
	\caption{ \textbf{Out of Domain Accuracy for Color-Mnist (VLAG Palette) in Figure \ref{fig:vlag-color-mnist-cont-overlap}}. Each sub-domain in Figure \ref{fig:vlag-color-mnist-cont-overlap} contains a random sample of 1000 mnist images. Random seed is shared for the different sub-domains of a nominal domain but different across nominal domains. Each repetition is with different starting random seed, 10 repetitions are done. The sub-domains are combined to form one nominal domain and 50 percent is used for training, the rest for validation. Comparison algorithms are DIVA \mcitep{ilse2019diva} and Match-DG \mcitep{mahajan2020domain}, while Deep-All is used as baseline by pooling all training domain s together. The 2nd domain is a bridge domain that connect the 1st and 3rd domain, so it is not used as test domain at all.  }
	\begin{tabular}{ p{2cm}cc }
		\toprule
		 \bfseries Color-Mnist (Figure \ref{fig:vlag-color-mnist-cont-overlap}) & \bfseries Test Domain 1& \bfseries Test Domain 3 \\
		\midrule
		DIVA &0.63 $\pm$ 0.05 &0.68 $\pm$ 0.03 \\
		
		HDUVA & \textbf{0.69} $\pm$ 0.05 & \textbf{0.71} $\pm$ 0.03 \\
		HDUVA\tiny{-no-zx} & 0.68 $\pm$ 0.04 &0.70 $\pm$ 0.02 \\
		Deep-All & 0.60 $\pm$ 0.05 & 0.68 $\pm$ 0.04 \\
		
		Match-DG  & 0.67 $\pm$ 0.06 &  0.70 $\pm$ 0.03\\
		\bottomrule
	\end{tabular}
	
	\label{tb:vlag-seq_color_overlap}
\end{table}
\begin{table}[!t]
	\centering
	\caption{ \textbf{Out of Domain Test Accuracy for Sequential Color-Mnist (Red Diverging Palette) from Figure \ref{fig:red-diverging-color-mnist-seq-overlap}}. Each sub-domain in Figure \ref{fig:red-diverging-color-mnist-seq-overlap} contains a random sample of 1000 mnist images. Random seed is shared for the different sub-domains of a nominal domain but different across nominal domains. Each repetition is with different starting random seed, 10 repetitions are done.  The sub-domains are combined to form one nominal domain and 50 percent is used for training, the rest for validation. Comparison algorithms are DIVA \mcitep{ilse2019diva} and Match-DG \mcitep{mahajan2020domain}, while Deep-All by pooling all training domains together is used as baseline. The 2nd domain is a bridge domain that connect the 1st and 3rd domain, so it is not used as test domain at all.}
	\begin{tabular}{ p{2cm}cc }
		\toprule
		\bfseries Color-Mnist (Figure \ref{fig:red-diverging-color-mnist-seq-overlap} 
		)  & \bfseries Test Domain 1 &  \bfseries Test Domain 3 \\
	    \midrule
		DIVA &0.53 $\pm$ 0.05 & 0.63 $\pm$ 0.05 \\
		\mname{} & \textbf{0.56} $\pm$ 0.05 & \textbf{0.68} $\pm$ 0.05 \\
		HDUVA\tiny{-no-zx} &0.55 $\pm$ 0.08 & 0.65 $\pm$ 0.04 \\
		Deep-All &0.53 $\pm$ 0.06 &0.61 $\pm$ 0.06 \\
		
		Match-DG  & 0.44 $\pm$ 0.04 &  0.67 $\pm$ 0.10\\
		\bottomrule

	\end{tabular}
	\label{tb:red-diverging-seq_color_overlap}
\end{table}




\subsection{Domain Generalization to Medical Image Classification}\label{sec:malaria}
Trustworthy prediction is essential for biomedical data where domain generalization poses a great challenge \mcitep{gossmann2019variational, gossmann2020performance,sun2019high}. For example, medical imaging datasets usually come from a multitude of patients and devices where both the patient and devices can form domains. In this study, as suggested by \mcitet{ilse2019diva}, we consider hospital as domains, which consist of patients as sub-domains. This correspond to hierarchical domains and  has practical implications. Since there can be thousands of patients, and having thousands of domain labels can be impractical and many patients can share common features, e.g. coming from nearby areas, but it can also be true that two hospitals can have similar patients. To simulate such a setting, we construct virtual hospitals by using the Malaria dataset~\mcitep{ilse2019diva, rajaraman2018pre} as described in Table \ref{tb:malaria}.  The Malaria dataset \mcitep{rajaraman2018pre} consist of thin blood smear slide images of
segmented cells from  Malaria patients. We group patients by their IDs for form hospitals. Table \ref{tb:malaria}
shows the out of domain classification accuracy across different algorithms. Our approach is able to implicitly learn the unobserved domain substructure of the data, resulting is substantially better accuracy on unseen test domains (i.e. a new hospital) compared to state-of-the-art approaches DIVA and MatchDG. The latter require explicit domain labels during training and fail to perform well in scenarios with domain substructure. Our approach also excels the standard Deep-All baseline where data across all domains is pooled together. 
\subsection{Domain Embedding}\label{sec:topic_plot}
\begin{figure}[!t]
	\centering
	\subfloat[test domain]{\includegraphics[width=0.17\textwidth]
		{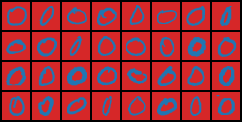}\label{fig:mnist-color0}}
	\subfloat[train\_domain1]{\includegraphics[width=0.17\textwidth]{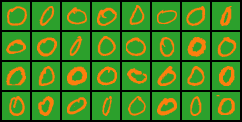}\label{fig:vlag-combo1}}
	\subfloat[train\_domain2]{\includegraphics[width=0.17\textwidth]{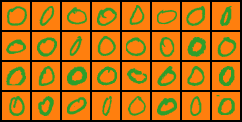}\label{fig:vlag-combo2}}
	\caption{\textbf{Color-Mnist} for domain embedding used as source data in Figure \ref{fig:topic_mnist_color}.} 
	\label{fig:color-mnist}
\end{figure}

\begin{figure}[h]
	\center\includegraphics[width=0.25\textwidth]{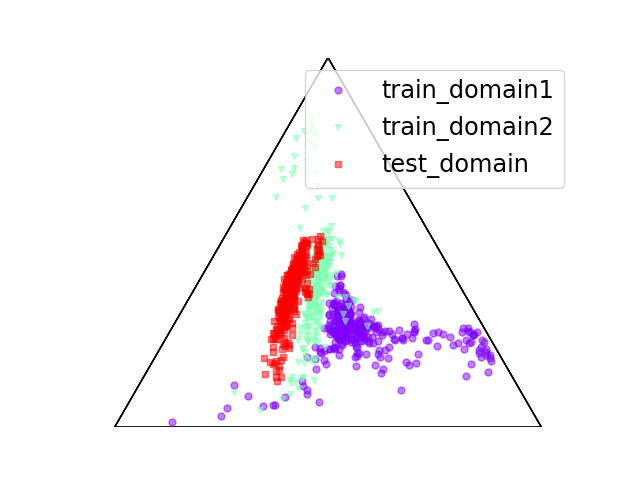}
	\caption{Topic plot of domains in Figure \ref{fig:color-mnist}}\label{fig:topic_mnist_color}
\end{figure}

Here, we investigate the ability of our approach to generate meaningful domain embeddings. We use the Color-Mnist in Figure \ref{fig:color-mnist} where the first domain is used as test domains and the 2nd and 3rd domain are used as training domains. 
Figure \ref{fig:topic_mnist_color} shows the topic plot of running HDUVA on this setting, where \mname{} assigns similar topics to  instances from the same domain, while samples from different domains can be well separated.
\subsection{State of the Art Domain Generalization  Benchmark}\label{sec:pacs}
We finally compare \mname{} to state-of-the-art domain generalization algorithms for a standard domain generalization task, where domain information is available on largely different domains. Table \ref{tb:pacs} shows algorithm performance on the PACS dataset \mcitep{li2017deeper} which is a popular domain generalization benchmark.~
We use AlexNet \mcitep{krizhevsky2012imagenet, krizhevsky2017imagenet} as the neural network architecture for  $q_{\phi_y}(z_y|x)$ cascaded with $q_{\omega}(y|z_y)]$ in our model in Equation \ref{eq:ehdiva_elbo_plus_y}. For fair comparison, the rest of the algorithms also use the same AlexNet as the classifier.

Table \ref{tb:pacs} shows that using the same AlexNet and further experimental settings detailed in \appname~\ref{supsec:hyper_exp_set}, the performance of \mname{} excels the state-of-the-art performances on the PACS dataset. 
Deep-All by pooling all training domain together remains as a strong baseline where we outperform Deep-All in 3 out of 4 test domains and ties the other one. Our algorithm has a better performance on the \texttt{Sketch} test domain with respect to DIVA. 
With the contrastive pretrain phase of \mcitet{mahajan2020domain} as the initialization for the AlexNet of \mname, the performance over the \texttt{Sketch} test domain further improves. 
Importantly, \mname{} achieves competitive performance without using domain labels during training and without using validation set (we conduct model selection using extended ELBO in Equation \ref{eq:ehdiva_elbo_plus_y}, see \appname~\ref{supsec:hyper_exp_set}). This enables domain generalization for a much wider range of use-cases than standard algorithms.
\begin{table}[t]
	\centering
	\caption{ \textbf {Malaria Virtual Hospital from Malaria Dataset}. Patients with ID starting with C1, C6, C8, C9 are grouped to form  4 virtual hospitals as 4 nominal domains. Virtual hospital C6 has 10 patients with 1061 infected cell images (in total 1748 images). Virtual hospital C8 has 10 patients with  957 infected cell images (in total 1638 images).  Virtual hospital C9 has 10 patients with 1284 infected cell images (in total 1964 images). Virtual hospital C1 has 90 patients with 8023 infected cell images (in total 14190 images). Each time, we combine the C6, C8, C9 virtual hospital domain as 3 training domains and sample 20 percent of the images for training. 20 random repetitions are done.  We report result on the test domain corresponding to virtual hospital C1. Comparison algorithms are DIVA \mcitep{ilse2019diva} and Match-DG \mcitep{mahajan2020domain}, while Deep-All is used as baseline by pooling all training domains together. \\Data source: \mcitep{rajaraman2018pre}}
		\begin{tabular}{lc}
		\toprule
		\bfseries Malaria Cell Classification  & \bfseries Test Accuracy \\
		\midrule 
		DIVA     &  0.83 $\pm$ 0.06\\
		\mname{}    & 0.87 $\pm$ 0.05 \\
		\mname{}-no-zx    & \textbf{0.88} $\pm$ 0.05 \\
		L\mname{} & 0.82 $\pm$ 0.06 \\
		Deep-All  &  0.84 $\pm$ 0.05 \\
		MatchDG   & 0.85 $\pm$ 0.09 \\
		\bottomrule
	\end{tabular}
	\label{tb:malaria}
\end{table}



\begin{table*}[t]
	\centering
	\small
	\begin{threeparttable}
		\caption[Domain Generalization in PACS Dataset with AlexNet]{\textbf{Domain Generalization in PACS Dataset with AlexNet}}
		\begin{tabular*}{\textwidth}{l @{\extracolsep{\fill}} lllll}
			\toprule
			\bfseries Methods 	      & \bfseries Art Painting &  \bfseries Cartoon & \bfseries	Photo & \bfseries Sketch & \bfseries  Ave.   \\
			\midrule
                  Jigsaw    &   \underline{48.05} $\pm$ 1.22           &  \underline{53.12} $\pm$ 1.50     &  \underline{67.39} $\pm$ 1.95              & 59.65 $\pm$ 1.20   &  \underline{59.55}   \\
                  Jigsaw(Caffe)$^{\ast}$    &  \tiny{\textit{64.20 $\pm$ 1.34}}            &  \tiny{\textit{66.08 $\pm$ 0.39}}     &  \tiny{\textit{89.51 $\pm$ 0.51}}              &  \tiny{\textit{62.32 $\pm$ 0.85}}  &  \tiny{\textit{70.53}}   \\
                  MatchDG     &60.69 $\pm$2.32 & 66.83 $\pm$ 1.54 & \textbf{87.96} $\pm$ 0.38 &\underline{55.60} $\pm$ 2.77 & 67.77\\
                   DIVA    & 64.61 $\pm$ 1.81    & 67.23 $\pm$ 1.25 &  87.35 $\pm$ 0.72 & 58.10 $\pm$ 2.65 & 69.32 \\
                     Deep-All & 64.35 $\pm$ 1.20 &  66.76 $\pm$ 2.28 &  85.25 $\pm$ 2.06  &  55.87 $\pm$ 2.63 & 68.06 \\
                  \mname{}  & \textbf{65.20} $\pm$ 1.17  & 66.40 $\pm$ 1.42 &  87.33 $\pm$ 0.77  & 61.04 $\pm$ 2.45 & 70.00 \\
                  W\mname{}$^{\ast\ast}$ &  64.21 $\pm$ 1.24  & \textbf{67.49} $\pm$ 0.70 & 87.48 $\pm$ 0.75 & 60.00 $\pm$ 1.97 & 69.80\\
                 
                 %
                  L\mname$^{\ast\ast\ast}$ & 65.29 $\pm$ 0.74  & 65.01 $\pm$ 2.15   & 87.11 $\pm$ 1.12 & 57.26 $\pm$ 2.02 & 68.67 \\
                  \mname-CTR$^{\ast\ast\ast\ast}$ & 65.12 $\pm$ 0.77  & 66.21 $\pm$ 1.70   & 87.50 $\pm$ 1.28 &  \textbf{63.10} $\pm$ 1.96 & \textbf{70.48} \\
                  \mname-no-zx$^{\ast\ast\ast\ast\ast}$ &65.15$\pm$ 0.99 & 66.63 $\pm$ 1.72 & 87.25 $\pm$ 1.43 & 60.77 $\pm$ 2.74 &69.95\\
			\bottomrule
		\end{tabular*}
		
		\begin{tablenotes}
		    \item See Section \ref{supsec:hyper_exp_set} for  experimental settings (hyper-parameters).  
			\item  
			$^{\ast}$ Jigsaw-Caffe: Using the author provided special weight initialization for Caffe-Alexnet instead of standard Pytorch Alexnet. So we do not consider this result for a fair comparison but just use this result to show the correctness of our implementation.\\
			$^{\ast\ast}$ W\mname: weak domain-supervision added to \mname{} as explained  in Appendix \ref{sec:weak}.\,\,\\ $^{\ast\ast\ast}$L\mname: Ladder \fullalgoname explained in Appendix \ref{sec:lhdiva}.\\
			$^{\ast\ast\ast\ast}$\mname-CTR: Use the contrastive learning phase (pretrain phase) of \mcitet{mahajan2020domain} as initialization for AlexNet of \mname.\\
			$^{\ast\ast\ast\ast\ast}$\mname-no-zx: \mname{} without $z_x$ variable, see Figure~\ref{fig:hduva_no_zx}.
		\end{tablenotes}
		\label{tb:pacs}
	\end{threeparttable}
\end{table*}


\subsection{Hyper-parameter sensitivity analysis}\label{subsec:hyper-sensitivity}
\begin{figure}
    \centering
    \includegraphics[width=0.5\textwidth]{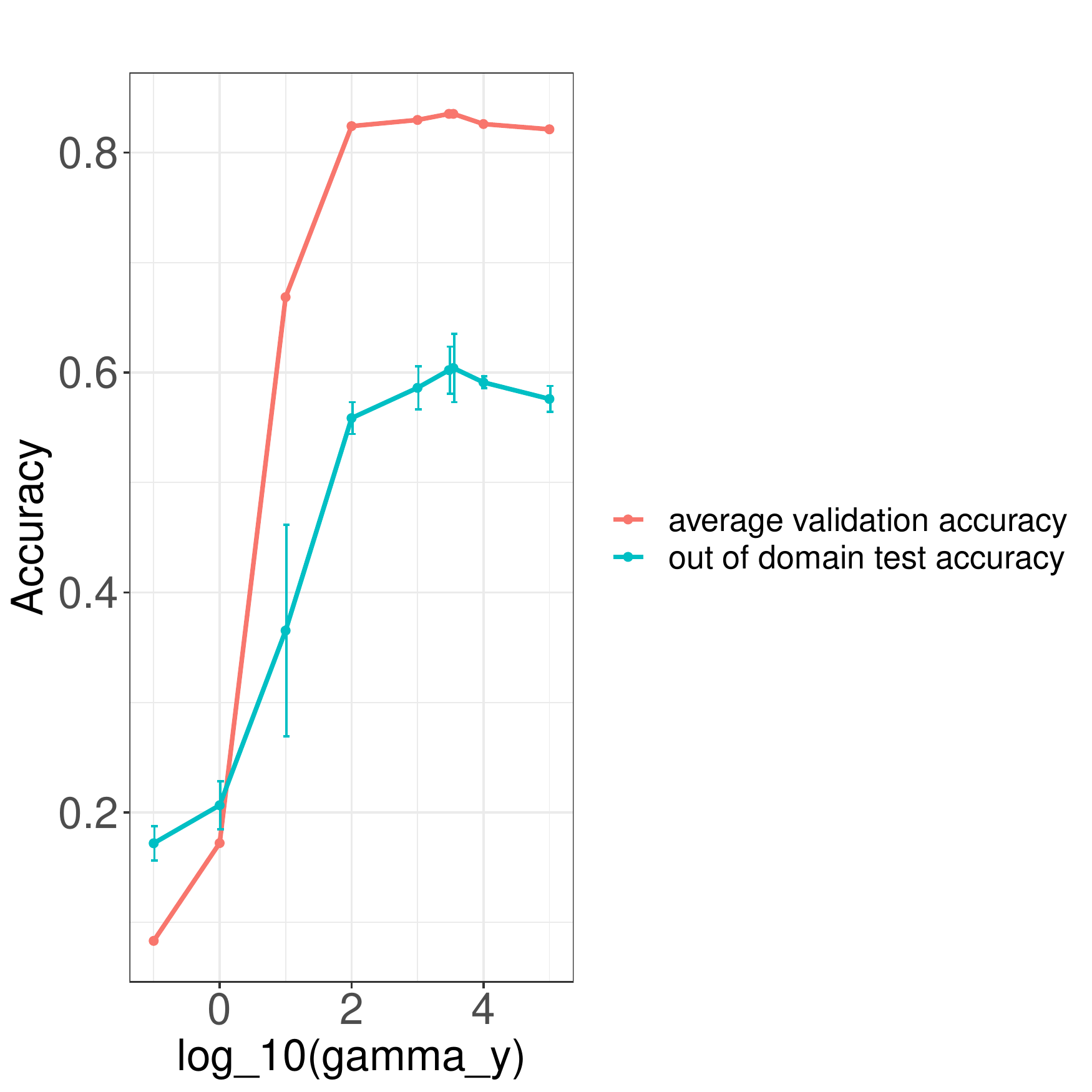}
    \caption{The effect of $\gamma_y$ in Equation~\ref{eq:ehdiva_elbo} on the performance of using \texttt{Sketch} as test domain from the PACS dataset. Results from 3 repetitions.}
    \label{fig:hyper_sensitivity}
\end{figure}

In Figure~\ref{fig:hyper_sensitivity}, we plot the effects of $\gamma_y$ in Equation~\ref{eq:elbo_plus_y} on the out-of-domain test accuracy of \texttt{Sketch} domain from the PACS dataset. Although with a fixed hyper-parameter, our algorithm does not need a validation set for early stopping, we do need a validation set to choose unknown hyper-parameters. From Figure~\ref{fig:hyper_sensitivity}, we can see that a well performed validation accuracy area corresponds to a well performed state-of-the-art out-of-domain test accuracy area. Note that in many domain generalization publications, when scanning hyper-parameters, they only give the out-of-domain test accuracy. Instead, here we argue that future authors should also report the validation accuracy indicating the hyper-parameter selection process, avoiding the risk of selecting the hyper-parameter in an oracle way.
\section{Conclusion}\label{sec:conclusion}
We proposed an Hierarchical Domain Invariant Variational Autoencoder, with the following improvements:
\begin{itemize}
\item Our approach does not require observed domain labels during training, facilitating domain generalization for a much wider range of applications. Additionally, our approach does not need validation set for model selection but only use extended ELBO for model selection.
\item In the presence of domain-substructure (hierarchical domain), our algorithm is able to robustly disentangle domain-specific variation from class-label specific variation. Besides, our algorithm is able to embed interpretable topics.
\item We proposed evaluation dataset for benchmarking hierarchical and sequential near continuous overlapped domain shift and showed that our algorithm could improve over competitor algorithms. 
\item Our algorithm has a competitive performance even in standard domain generalization tasks, where observed domain information is available on clearly separated domains. 
\end{itemize}
\appendix

\begin{center}
\textbf{\large Appendix}
\end{center}
\appendix
In section \ref{sec:no-zx}, we introduce an alternative model without $z_x$ variable for \mname{}.
In section \ref{sec:lhdiva} we explain an alternative inference algorithm inspired by Ladder-VAE   \mcitep{sonderby2016ladder} for our proposed model. In section \ref{sec:weak}, we introduce weak domain supervision methods for both inference algorithms. In section \ref{supsec:hyper_exp_set}, we list further details on experimental settings.
\section{Model Ablation: Graphical Model of \mname{} without $z_x$}\label{sec:no-zx}
\begin{figure}[H]
\centering
\includegraphics[width=0.3\textwidth]{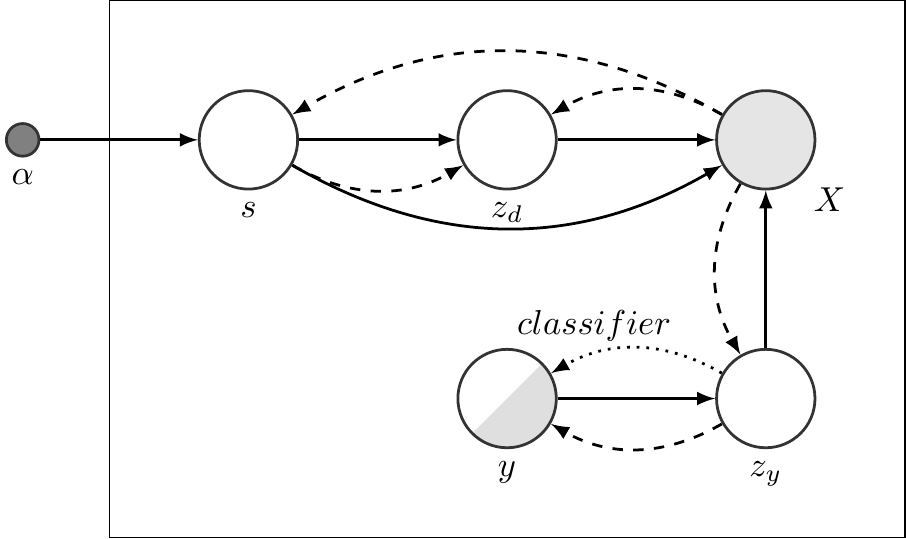}
\caption{HDUVA without variable $z_x$}\label{fig:hduva_no_zx}
\end{figure}

\section{Model Ablation: Alternative Inference Method for \mname}\label{sec:lhdiva}
 
\begin{figure}[h!]
	\center\includegraphics[width=0.4\textwidth]{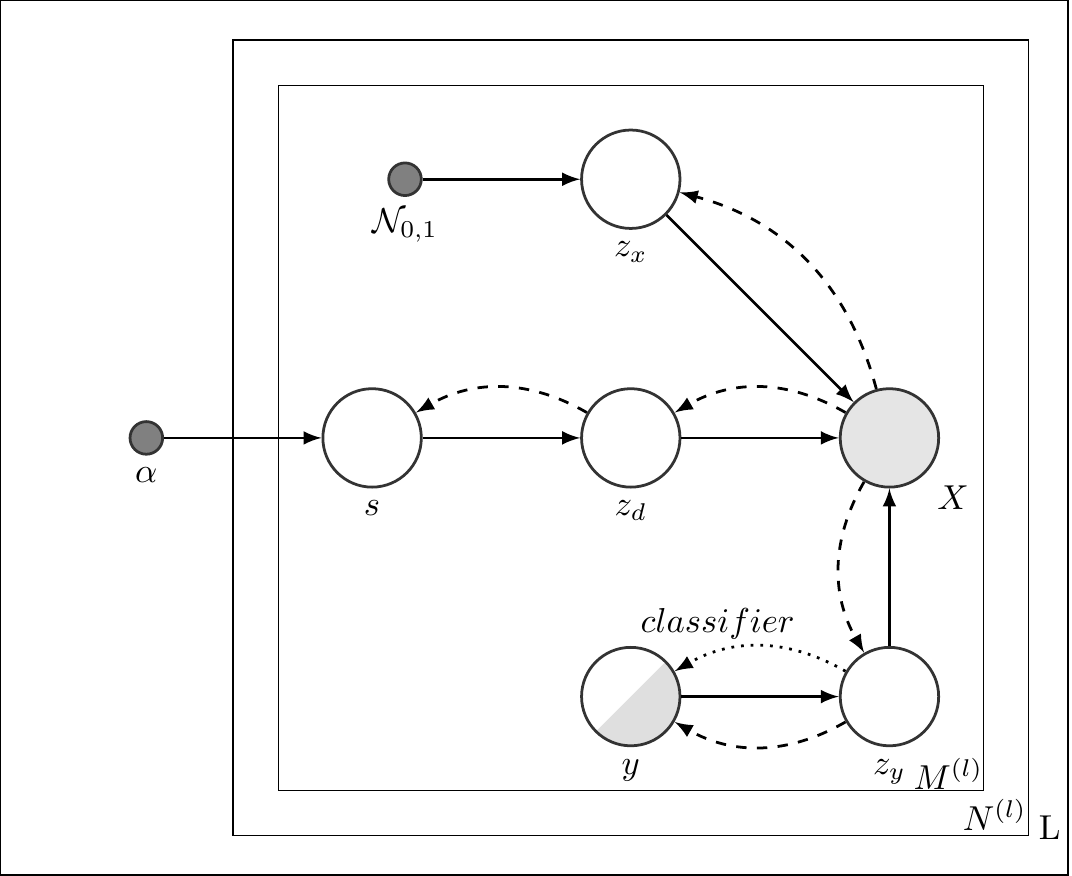}
	\caption{L\mname: Ladder \fullalgoname}
	\label{fig:lhdiva}
\end{figure}

 We propose an alternative inference algorithm for our model. 
 The graphical model for the Ladder-VAE version of our model is shown in Figure \ref{fig:lhdiva} which we coined L\mname. The corresponding variational posterior and ELBO is explained below. 
 We summarize this alternative  algorithm in Algorithm \ref{algo:hdiva}.
 
\subsection{Inference for \lname{}}
In Figure \ref{fig:lhdiva}, we factorize the approximate posterior as follows:
\begin{align}
&q_{\phi}(s^{(l,i)}, z_d^{(l,i)}, z_x^{(l,i)}, z_y^{(l,i)}|x^{(l,i)})\nonumber\\
=&q_{\phi}(s|z_d^{(l,i)})q_{\phi}(z_d^{(l,i)}, z_x^{(l,i)}, z_y^{(l,i)}|x^{(l,i)})  \label{eq:qs_qzd_qzx_qzy}
\end{align}

For the approximate posterior distributions of $z_x$, $z_d$ and $z_y$, we follow \mcitet{ilse2019diva} and assume fully factorized Gaussians with parameters given as a function of their input:

\begin{align}
&q_{\phi}(z_d^{(l,i)}, z_x^{(l,i)}, z_y^{(l,i)}|x^{(l,i)})\nonumber\\
=&q_{\phi_d}(z_d^{(l,i)}|x^{(l,i)})q_{\phi_x}(z_x^{(l,i)}|x^{(l,i)})q_{\phi_y}(z_y^{(l,i)}|x^{(l,i)})  \label{eq:qzd_qzx_qzy}   
\end{align}
Encoders $q_{\phi_y}$, $q_{\phi_d}$ and $q_{\phi_x}$ are parameterized by $\phi_y$, $\phi_d$ and $\phi_x$ using separate neural networks to model respective means and variances as function of $x$.



For the form of the approximate posterior distribution of the topic $s$ we chose a Dirichlet distribution:
\begin{align}
&  q_{\phi_s}(s^{(l,i)}|z^{(l,i)}_{d_{j}})
= Dir\left(s^{(l,i)}| \phi_{s}(z^{(l,i)}_{d_{j}})\right)  \label{eq:alpha_infer_topic}\end{align}
where $\phi_{s}$ parameterizes the concentration parameter based on $z_d$, using a neural network.
\subsection{ELBO for \lname{}}
Given the priors and factorization described above, we can optimize the model parameters by maximizing the evidence lower bound (ELBO). 
We can write the ELBO for a given input-output tuple $(x,y)$ as:
\begin{align}
&ELBO(x,y) = E_{q(z_d|x), q(z_x|x), q(z_y|x)}\log p_{\theta}(x|z_d, z_x, z_y) \nonumber\\
&- \beta_x KL(q_{\phi_x}(z_x|x)||p_{\theta_x}(z_x)) - \beta_y KL(q_{\phi_y}(z_y|x)||p_{\theta_y}\nonumber\\
&(z_y|y)) - \beta_d 
E_{q_{\phi_s}(s|x,z_d), q_{\phi_d}(z_d|x)} \log \frac{q_{\phi_d}(z_d|x)}{p_{\theta_d}(z_d|s)} \nonumber\\ &-\beta_sE_{q_{\phi_d}(z_d|x)}KL(q_{\phi_s}(s|z_d)||p_{\theta_s}(s|\alpha)) \label{eq:elbo}
\end{align}
where we use $\beta$ to represent the multiplier in the Beta-VAE setting \mcitep{higgins2016beta}, further encouraging disentanglement of the latent representations.

We add an auxiliary classifier $q_{\omega}(y|z)$, which is parameterized by $\omega$, to encourage separation of classes $y$ in $z_y$. The L\mname{} objective then becomes:
\begin{align}
\mathcal{F}(x,y) = ELBO(x,y) + \gamma_y E_{q_{\phi_y}(z_y|x)}[\log q_{\omega}(y|z_y)]\label{eq:elbo_plus_y}
\end{align}

To efficiently perform inference with the dependent stochastic variables $z_d$ and $s$, we follow \mcitet{sonderby2016ladder} and adapt the ELBO using the Ladder VAE approach as detailed in the next section.

\subsubsection{Dealing with Dependent Stochastic Variables} \label{para:hierarchy}
The joint posterior $q(z_d,s|x)$ can be written as:
\begin{align}
q(z_d,s|x) &= \frac{q(z_d,s,x)}{q(x)} =\frac{q(z_d,s,x)}{q(z_d,x)}\frac{q(z_d,x)}{q(x)}\nonumber \\&= q(s|z_d,x)q(z_d|x) = q(s|z_d)q(z_d|x)\label{eq:cond_independence}
\end{align} 
where conditional independence of $s$ from $x$ is assumed.
As pointed out by \mcitet{chen2016variational, tomczak2018vae}, this can lead to inactive stochastic units. We follow \mcitet{sonderby2016ladder} and recursively correct the generative distribution by a data dependent approximate likelihood.
Additionally, we implement a deterministic warm-up period of $\beta$ following \mcitet{sonderby2016ladder, ilse2019diva}, in order to prevent the posterior of the latent representation from aligning too quickly to its prior distribution.
\section{Weak Supervision on domains}\label{sec:weak}
In many scenarios only incomplete domain information is available. For example, due to privacy concerns, data from from different customers within a region may be pooled so that information on the nominal domain at customer-level is lost and only higher-level domain information is available. In other settings, substantial heterogeneity may exist in a domain and various unobserved sub-domains may be present. We introduce two techniques for weak supervision on domains, allowing the model to infer such lower-level domains or sub-domain information in the form of a topic $s$.

\subsection{Topic Distribution Aggregation}\label{sec:agg_alpha} To indicate that a group of samples "weakly" belong to one domain, we aggregate the concentration parameter of the posterior distribution of $s$ for all samples in a minibatch (note that all samples in a minibatch have the same nominal domain):  
\begin{align}
& \phi^{agg}_{s}(z^{(l,i)}_{d_{1:M}}) = 1/M\sum_{j=1:M}\left(\phi_{s}(z_{d_j}^{(l,i)})\right) \label{eq:agg_alpha}
\end{align}

We then use the aggregated concentration parameter to sample a topic from a Dirichlet distribution:
\begin{align}&  q^{agg}(s^{(l,i)}|z^{(l,i)}_{d_{1:M}})= Dir\left(\cdot| \phi_{s}^{agg}(z^{(l,i)}_{d_{1:M}})\right) \label{eq:alpha_infer_agg}\end{align}
The conditional prior of $z_d^{(l,i)}$ (equation \ref{eq:prior_zd}) then shares this same topic for all samples in the $i$th mini-batch. We interpret this topic-sharing across samples in a mini-batch as a form of regularized weak supervision. In one-hot encoded approaches, all samples from the same nominal domain would share the same topic. In contrast, sharing a topic in the conditional prior of the latent representation across samples in a mini-batch provides a weak supervision, whilst allowing for an efficient optimisation via SGD.
Note that concentration parameters for a mini-batch are only aggregated during training, at test time sample-specific posterior concentration parameters are used. 

\subsection{Weak domain distribution supervision with MMD}\label{sec:mmd}
DIVA encourages separation of nominal domains in the latent space $z_d$ by fitting an explicit domain classifier which might limit model performance in the case of incomplete domain information. To mitigate these limitations  but still weakly encourage separation between different nominal domains, we constrain the \mname{} objective based on the Maximum-Mean-Discrepancy (MMD) \mcitep{gretton2012kernel} between pairwise domains. 

Denoting $C_{mmd}^d$ as the minimal distance computed by MMD as an inequality constraint, we can write the constraint optimization of equation \ref{eq:elbo_plus_y} as follows: 
\begin{align}
&\underset{\theta, \phi, \omega}{argmax}\,\sum_{l,i}\mathcal{F}(x^{(l,i)}, y^{(l,i)})\nonumber\\
&s.t. \,\,\,  MMD(q_{z_d}^{(l,i)}(\cdot)|q_{z_d}^{(l^{'},i)}(\cdot)) \ge C_{mmd}^{(l, l^{'})}\label{eq:mmd}
\end{align}

\subsection{Practical considerations}
In practice, we transform the constrained optimization in Equation \ref{eq:mmd} with a Langrange Multiplier. This leads to the final loss in Equation \ref{eq:final}, where $\gamma_d^{(l)}$ denotes the Lagrange multiplier for $C_{mmd}^d$ (c.f. Equation \ref{eq:mmd}):
\begin{align}
\mathcal{L} = &\sum_{l,i}-\mathcal{F}^{(agg, ladder)}(x^{(l,i)},y^{(l,i)}) \nonumber\\ &-\gamma_d^{(l)}\sum_{i, l,l^{'}}MMD(q_{z_d}^{(l,i)}(\cdot)|q_{z_d}^{(l^{'},i)(\cdot)}) 
\label{eq:final}\end{align}
Superscript $agg$ and $ladder$ in Equation \ref{eq:final} refer to batch-wise aggregation of the concentration parameter and the ladder approach described above.

\begin{algorithm}
	\caption{L\mname{}} 
        \begin{algorithmic}[1] 
		\WHILE{not converged or maximum epochs not reached}
		\STATE warm up $\beta$ defined in Equation \ref{eq:elbo}, as in \mcitep{sonderby2016ladder}
		\STATE fetch mini-batch \{x, y\} =\{$x^{(l,i)},y^{(l,i)}$\}  \\
		\STATE compute parameters for $q_{\phi_x}(z_x|x)$, $q_{\phi_y}(z_y|x)$, $q_{\phi_d}(z_d|x)$ \\
		\STATE sample latent variable $z_{x}^q$, $z_{y}^q$ and compute $[\log q_{\omega}(y|z_y)]$ in equation \ref{eq:elbo_plus_y}
		\STATE sample $z_d^q$, infer concentration parameter
		$\phi_{s}(z_d)$ and aggregate according to Equation \ref{eq:agg_alpha}
		\STATE sample topic $s$ from aggregated
		 $\phi^{agg}_{s}(z_{d_{1:M}})$ according to Equation \ref{eq:alpha_infer_agg}.
		\STATE compute prior distribution for $z_d$ using $s$
		\STATE adapt posterior of $q_{\phi_d}(z_d)$ with ladder-vae method \mcitep{sonderby2016ladder}
		\STATE sample $z_{d}^q$ from adapted $q_{\phi_d}(z_d)$ 
		\STATE compute $p_{\theta}(x|z_x,z_y,z_d)$ using sampled $z_{x}^q$, $z_{y}^q$, $z_{d}^q$
		\STATE compute KL divergence for $z_d$, $z_x$ and $z_y$, $s$ in Equation \ref{eq:elbo}
		\STATE compute pair wise MMD of the nominal domains 
		\STATE aggregate loss according to \ref{eq:final} and update model
		\ENDWHILE 
	\end{algorithmic}
	\label{algo:hdiva}
\end{algorithm}

\section{Other experiment details}\label{supsec:hyper_exp_set}
For comparing algorithms, we implemented DIVA \mcitep{ilse2019diva} and MatchDG \mcitep{mahajan2020domain}, and use the same hyper-parameters suggested by the original paper. 
For \mname{}, we match the hyper-parameters in \mcitep{ilse2019diva}, where we take the latent dimension for each latent code is taken to be 64, i.e. $z_x=z_y=z_d=64$.
The classifier is taken to be a one layer neural network with Relu activation. For all experiments, 
$\gamma_y$ in equation \ref{eq:ehdiva_elbo_plus_y} is taken to be $1e5$, while the $\beta$ values are taken to be 1, warm-up of KL divergence loss in Equation \ref{eq:ehdiva_elbo} is taken to be 100 epochs.
We use topic dimension of 3 for \mname{}.
For the malaria experiment, we run with maximum 1000 epochs, with early stopping tolerance of 100 epochs. For \mname{}, we use ELBO directly as model selection criteria, for the rest of the algorithms, we use validation accuracy as model selection criteria. That means, we do not use the validation set at all. 

The mnist related experiments are run with maximum 500 epochs with early stopping tolerance of 100 epochs. For \mname{}, we use ELBO directly as model selection criteria, for the rest of the algorithms, we use validation accuracy as model selection criteria. That means, we do not use the validation set at all.
We use a learning rate of 1e-4 for DIVA and \mname{}, a learning rate of 1e-5 (better than 1e-4) for Deep-All and the suggested learning rate for MatchDG.
For all experiments regarding MNIST, 
we use random sub-samples (each contains 1000 instances) pre-sampled from \url{https://github.com/AMLab-Amsterdam/DIVA/tree/master/paper_experiments/rotated_mnist/dataset} with commit hash tag
\url{ab590b4c95b5f667e7b5a7730a797356d124}.

For the PACS experiment, 
we use Pytorch \texttt{torchvision} with version 0.8.2 for AlexNet initialization using\begin{verbatim}
 torchvisionmodels.alexnet(pretrained=True)  
\end{verbatim} 
We run with maximum 500 epochs, with early stopping criteria of 5 epochs to save computation resources. However, we found that DeepAll with early-stop tolerance of 5 epochs leads to extremely bad results, so we use 100 epochs as early stop tolerance for DeepAll, which forms a strong baseline.
For \mname{}, we use ELBO directly as model selection criteria, for the rest of the algorithms, we use validation accuracy as model selection criteria. That means, we do not use the validation set at all.
We use a learning rate of 1e-5 for \mname{}, DIVA, Deep-All. For MatchDG, we implemented the algorithm by refactoring the author's commit \url{https://github.com/microsoft/robustdg/commit/8326ac96ac3b0062c625669c8941b29c71c5f0ad} which we also use for the CTR version of our algorithm. 
We use the suggested optimal hyper-parameters according to Table 7 of the MatchDG paper in \citet{mahajan2020domain}, which is learning rate 0.001, weight decay of 0.001. 
For Jigsaw, we use the default permutations number of 31 as suggested by the code of the authors for PACS from \url{https://github.com/fmcarlucci/JigenDG/commit/8522c291233588b72aa689f3665a8c14f0ab1a74}. As suggested in the code, the jigsaw loss weight is taken to be 0.9. However, we found that using the standard AlexNet from Pytorch, when setting learning rate as 0.001 lead to extremely bad results for Jigsaw, so we use the learning rate of 1e-5 as other algorithms. 


\subsection{Architectures}\label{sec:architecture}
To make a fair comparison of the competing algorithms, we use the same neural network architectures for all algorithms being compared.
We list the neural network architectures used in the experiment as follows: For the decoder, we use the same architecture as in Table~\ref{tab:decoder} for both DIVA and \mname{} across the experiments.
\begin{table}[h]
\begin{center}
\caption{Decoder Architecture\\We use the same decoder architecture across all experiments. First, from latent code, we use a Gated Dense Layer \citep{tomczak2018vae} to map the latent code to the dimension of the image. Then we apply two times Gated Convolution\cite{pixelcnn}(kernel size =3, stride=1, padding=1, dilation=1)).}\label{tab:decoder}
	\begin{tabular}{| c c c c |}
		 \hline
		  & Block      & Input  & Output\\
		1 & GatedDense & dim(z) &3 $\times$ height $\times$ width \\
		2 & GatedConv2d & 3 &64 \\
		3 & GatedConv2d & 64 &64 \\
		 \hline	 
	\end{tabular}
\end{center}
\end{table}
For the encoder, for image of size 224, we used Alexnet with the last layer removed as the encoder.
For the synthetic datasets as well as the malaria dataset, we use the architecture as described in Table~\ref{tb:encoder}.
\begin{table}[h]
\begin{center}
  \caption{Encoder architecture  used for small images. The following architecture is with convolution kernel size = 5, convolution stride size = 1, max pool stride size =2. \label{tb:encoder}}
	\begin{tabular}{| c c c c |}
		 \hline
		  & Block      & Input  & Output\\
		1 & Conv2d & 3 &32 \\
		2 & BatchNorm2d & 32 & 32 \\
		3 & Relu &  & \\
                4 & MaxPool2d(2) &  & \\
                4 & Conv2d & 32 & 64 \\
		5 & BatchNorm2d & 64 & 64 \\
                6 & ReLu &  & \\
                7 & MaxPool2d(2) &  & \\
		 \hline	 
	\end{tabular}
\end{center}
\end{table}



\bibliography{dg}
\end{document}